\documentclass{article}
\usepackage{ijcai18}

\usepackage{times}
\usepackage{xcolor}
\usepackage{soul}
\usepackage{amsmath}
\usepackage{amsfonts}
\usepackage{url}
\usepackage{graphicx}
\usepackage{booktabs}
\usepackage{array}
\usepackage{enumitem}
\usepackage{titlesec}
\usepackage[utf8]{inputenc}
\usepackage[small]{caption}
\DeclareMathOperator*{\argmin}{arg\,min}

\usepackage{subcaption}
\title{DRPose3D: Depth Ranking in 3D Human Pose Estimation}

\author{
Min Wang$^1$,
Xipeng Chen$^2$,
Wentao Liu$^3$,
Chen Qian$^4$,
Liang Lin$^{2,4}$,
Lizhuang Ma$^{5,1}$
\\
$^1$ Department of Computer Science and Engineering, Shanghai Jiao Tong University \\
$^2$ School of Data and Computer Science, Sun Yat-Sen University \\
$^3$ Department of Computer Science and Technology, Tsinghua University \\
$^4$ SenseTime Group Limited\\
$^5$ School of Computer Science and Software Engineering, East China Normal University\\
 yinger650@sjtu.edu.cn,
 chenxp37@mail2.sysu.edu.cn,
 liuwtwinter@gmail.com \\
 qianchen@sensetime.com,
 linliang@ieee.org,
 ma-lz@cs.sjtu.edu.cn\\
}
\begin{document}

\maketitle

\begin{abstract}
In this paper, we propose a two-stage depth ranking based method (DRPose3D) to tackle the problem of 3D human pose estimation. Instead of accurate 3D positions, the depth ranking can be identified by human intuitively and learned using the deep neural network more easily by solving classification problems. Moreover, depth ranking contains rich 3D information. It prevents the 2D-to-3D pose regression in two-stage methods from being ill-posed. In our method, firstly, we design a Pairwise Ranking Convolutional Neural Network (PRCNN) to extract depth rankings of human joints from images.  Secondly, a coarse-to-fine 3D Pose Network(DPNet) is proposed to estimate 3D poses from both depth rankings and 2D human joint locations. Additionally, to improve the generality of our model, we introduce a statistical method to augment depth rankings. Our approach outperforms the state-of-the-art methods in the Human3.6M benchmark for all three testing protocols, indicating that depth ranking is an essential geometric feature which can be learned to improve the 3D pose estimation.
\end{abstract}

\section{Introduction}
3D human pose estimation is an important problem that relates to a variety of applications such as human-computer interaction, augmented reality and behavior analysis, etc.
Unlike 2D human pose dataset\cite{andriluka20142d}, in which the ground-truth can be obtained from manual labeling, the 3D pose is hard to get without sophisticated tracking devices.

In 3D human pose estimation, end-to-end methods \cite{pavlakos2017coarse,Sun_2017_ICCV} map input images to 3D joint positions directly. Their advantages lie in the ability to use shading, occlusion, appearance information contained in images. However, these images are captured in the laboratory environment such as Human3.6M \cite{ionescu2014human3} and data augmentation for images can hardly be performed in 3D space. On the other hand, two-stage methods with a simple model \cite{martinez2017simple} have achieved competitive performance on the Human3.6M dataset. These methods first predict the 2D locations of the human joints, then estimate 3D poses based on their 2D joint predictions. The first stage can utilize adequate 2D human pose datasets. The second stage can use data augmentation to simulate 2D and 3D data in 3D space to fully utilize the Human3.6M dataset. However, rich 3D information contained in images is not used. Besides, estimating 3D positions from 2D locations in the second stage is an ill-posed problem since multiple 3D poses can have the same 2D projection.

\begin{figure}[t]
	\setlength{\abovecaptionskip}{0.2cm}	
	\setlength{\belowcaptionskip}{-0.4cm}
	\centering
	\includegraphics[width = .5\textwidth]{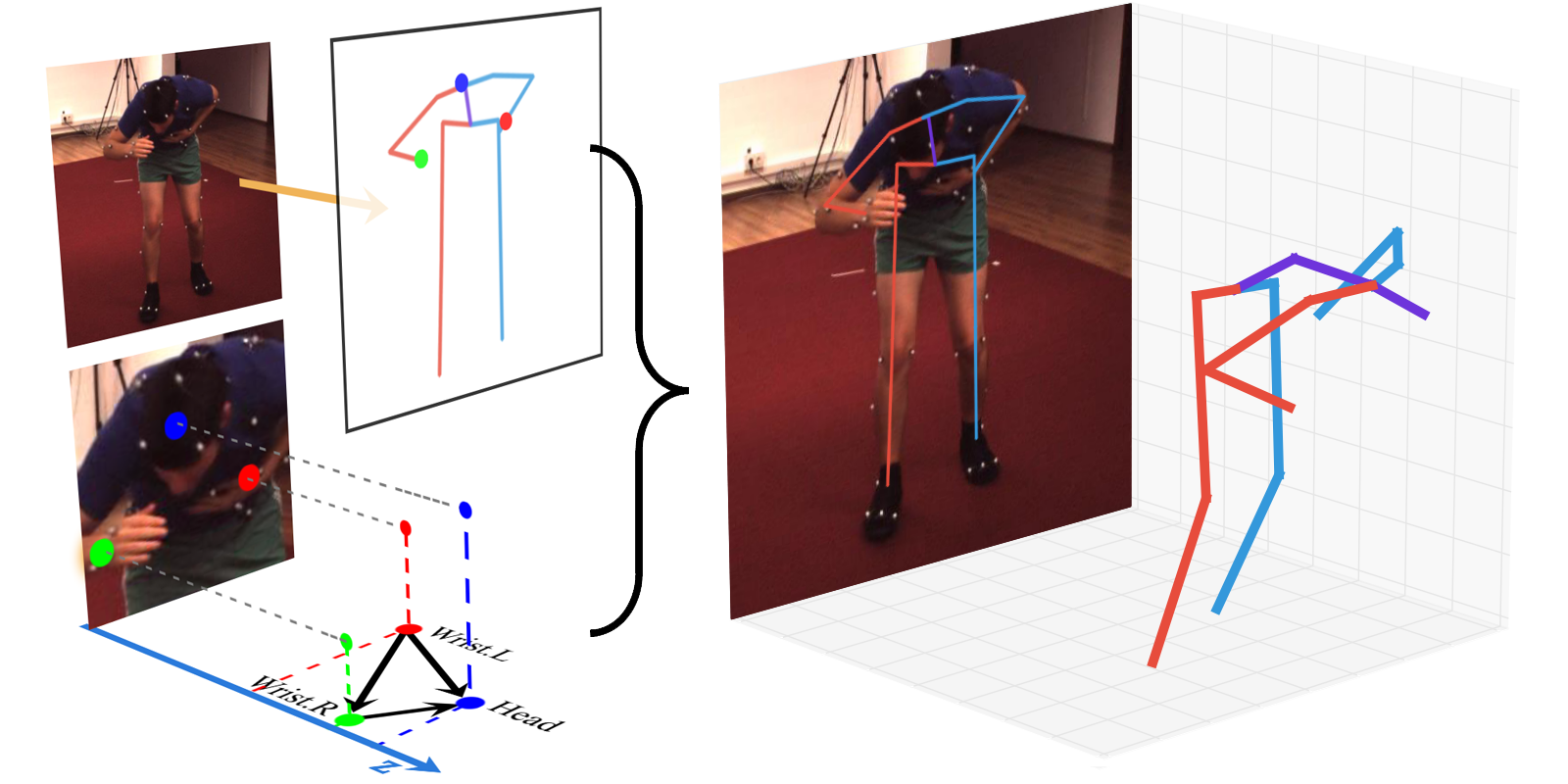}
	\caption{
		Illustrations on how depth ranking works in 3D pose estimation.
		We estimate 2D joint locations and depth rankings between joints from a single image (shown on the left) and combine them together to obtain 3D poses. The pairwise rankings are represented by a directed graph, where the edge points from the behind to the front, e.g., the right wrist is in front of the left wrist on the z-axis.	
	}
	\label{fig:motivation}
\end{figure}

Depth ranking encodes rich 3D information. As mentioned in Section.3, under some weak assumptions, it can uniquely decide 3D pose when combined with the 2D pose. It changes the second stage, a 2D-to-3D problem in previous methods, to a well-defined one. Besides, as a geometric property, depth ranking can also be augmented in 3D space. As the second stage being a one-to-one function with depth ranking, data augmentation can be performed fully without concerns.  Moreover, for depth ranking, we learn the relationship between each pair of joints, for example in Figure~\ref{fig:motivation}, the wrist lies before the head. This makes depth ranking problem a series of classification problems, which can be effectively solved by deep neural networks.

To this end, we propose a two-stage method called Depth Ranking 3D Human Pose Estimator (DRPose3D). In Figure~\ref{fig:framework}, our method is divided into two stages. It explicitly learns depth rankings between each pair of human joints from images and then uses it together with 2D joint locations to estimate 3D poses. We make three contributions to investigate the utility of depth rankings in 3D human pose estimation.

\begin{itemize}[leftmargin=*]
	\item We design a \textit{Pairwise Ranking Convolutional Neural Network (PRCNN)} to extract the depth rankings of pairwise human joints from a single RGB image. PRCNN transforms the depth ranking problem into the pairwise classification problem by generating a ranking matrix representing the depth relations between each pair of human joints.
	
	\item We propose a coarse-to-fine 3D Pose Estimator named \textit{DPNet} composed of the \textit{DepthNet} and the \textit{PoseNet}. It regresses the 3D pose from 2D joint locations and the depth ranking matrix. Since noises exist in the estimated ranking matrix, directly using the ranking matrix and 2D joint locations will lead to poor performance. \textit{DPNet} first estimates coarse depth value that is consistent with majority entries of the depth ranking matrix then regresses the accurate 3D poses in a coarse-to-fine manner.
	
	\item
	Data augmentation in 3D space for the second stage is explored. By synthesizing 3D poses and camera parameters, 2D poses and ranking matrices can be generated adequately. Unlike previous work, synthesized cameras are put around the same circle, which is unknown in real scenarios. We randomly sample camera positions on all possible positions around the subject. To make the augmented data obey the data distribution of training dataset, we use a statistical method to add noises.
\end{itemize}

The proposed DRPose3D framework achieves the-state-of-the-art results on three common protocols of Human3.6M dataset compared with both end-to-end and two-stage methods \cite{Sun_2017_ICCV,fang2017learning,martinez2017simple}.
\textit{Mean per joint position errors (MPJPE)} on the three protocols are decreased to $57.8mm (2.2\%\downarrow)$, $42.9mm (6.1\%\downarrow)$ and $62.8mm (13.7\%\downarrow)$ respectively. And the MPJPE gap between protocol \#3 and protocol \#1 is reduced to $5.0mm(59.7\%\downarrow)$. It proves that our method is robust to new camera positions and our data augmentation is very effective. The experimental results show that the depth ranking is an essential geometric knowledge that can be learned, utilized and augmented in 3D pose estimation.

\section{Related Work}
\paragraph{Learning to Rank}

Learning to rank is widely used in computer science tasks, especially in information retrieval systems. A lot of methods have been proposed in the literature, i.e., point-wise, pairwise and list-wise \cite{cao2007learning}. Among these methods, the pairwise ranking is the most popular because of more efficient data labeling.
\cite{cohen1998learning} performs a two-stage framework that exploits a preference function to get pairwise rankings.
RankNet \cite{burges2005learning} proposes gradient descent to learn a binary classifier which indicates the pairwise ranks. Their methods provide an effective way of ranking learning but have to extract hand designed features for each item.
With the great advance of deep learning, there are increasing applications of rankings such as age estimation\cite{chen2017using} and face beautification\cite{li2015deep}. These methods try to learn rankings with neural networks but focus on image level attributes. \cite{chen2016single} proposes a pairwise ranking method for depth estimation. However, it only learns to rank one pair of pixels in an image explicitly. Different from previous methods, PRCNN learns the depth ranking between each pair of human joints implicitly.

\paragraph{3D Pose Estimation}
3D pose estimation methods can be divided into two types: the end-to-end methods and the two-stage methods.
\textit{End-to-end method} in 3D pose estimation benefits from the completeness of image information but suffers from hardness of accurate 3D localization and limited 3D human pose datasets.
In order to better locate 3D joint positions, \cite{Sun_2017_ICCV} proposes to regress bone based representation instead of joints. \cite{pavlakos2017coarse} uses a volumetric representation to estimate 3D pose in a coarse-to-fine manner. These methods  still need 2D image and 3D pose pairs. \cite{rogez2016mocap} augments images by assembling multiple image fragments according to 3D poses, but the synthetic images contain a lot of artifacts.
\cite{mehta2016monocular} transfers 2D pose detector into a 3D regression network and \cite{zhou2017towards} learns 3D human poses with extra 2D annotated images. These methods benefit from the diversity of 2D human pose datasets.

\textit{Two-stage methods} usually predict 2D poses first, then use optimization or machine learning methods to obtain 3D pose results.
\cite{moreno20173d} converts 2D and 3D pose data into an Euclidean distance matrix. It uses CNN to regress 3D pose distance matrix from 2D pose distance matrix.
\cite{martinez2017simple} established a simple baseline by regressing 3D pose directly from 2D joint locations.
\cite{fang2017learning}, \cite{akhter2015pose} encode prior human body shape constrains into 2D-to-3D pose estimation.
Some other methods propose to search approximate poses from a designed 3D pose library \cite{mori2006recovering,ramakrishna2012reconstructing,bogo2016keep,jahangiri2017generating,linCVPR17RPSM}. These methods focus on inferencing possible 3D poses from human body constrains and ignore other geometric knowledge embedded in the image features.

\section{Depth Ranking in 3D Pose Estimation}

Depth ranking is an important cue to infer the 3D joint positions. Recovering 3D poses purely from 2D joint locations is an ill-posed problem. The use of depth ranking would alleviate the ill-posedness. We represent the 2D skeleton by $S_{2D} = ((x_1, y_1), ..., (x_n, y_n)) \in\mathbb{R}^{2n}$. Given z-axis pointing towards the screen, the 3D pose $S_{3D} =((x'_1, y'_1, z'_1), ..., (x'_n, y'_n, z'_n))\in\mathbb{R}^{3n}$. Under the assumption of orthogonal projection and fixed length $l$ between two adjacent joints, $i$ and $j$, we have
$|z'_i-z'_j| = \sqrt{l^2-(x'_i-x'_j)^2-(y'_i-y'_j)^2}$.
If the depth ranking between adjacent joints is known, the relative 3D position between joint $i$ and joint $j$ is determined.
Thus with the knowledge of depth rankings between adjacent joints, 2D joint locations, and limb length priors, the 3D skeleton is almost determined.

\begin{figure}[t]
	\setlength{\abovecaptionskip}{0.2cm}	
	\setlength{\belowcaptionskip}{-0.35cm}
	\centering
	\includegraphics[width = 0.5\textwidth]{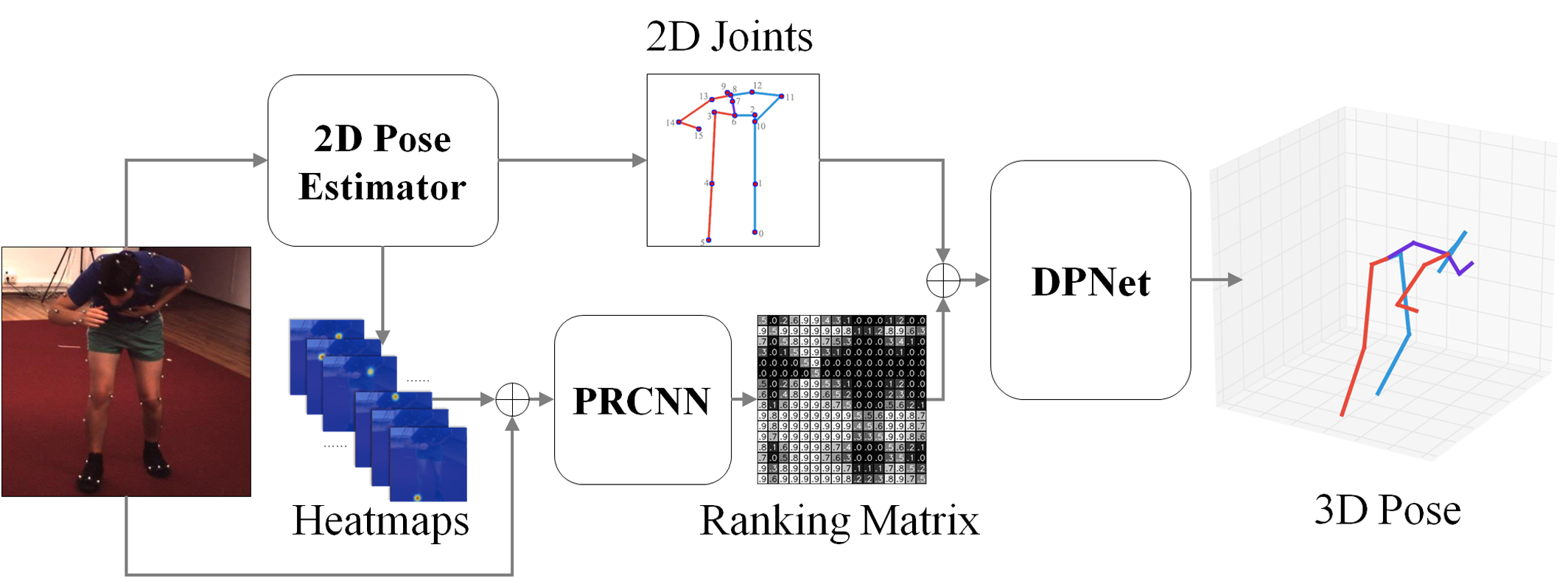}
	\caption{
		An overview of DRPose3D. First we use a 2D pose estimator to generate the joint heatmaps from an image.
		Then we concatenate the heatmaps and the original image as input for PRCNN to predict the pairwise ranking matrix.
		Finally, DPNet regresses 3D pose from the 2D joint locations and the pairwise ranking matrix.				
	}
	\label{fig:framework}
\end{figure}
In order to learn depth ranking effectively, we introduce a pairwise ranking matrix to represent the depth ranking. By using pairwise ranking matrix, we transform the depth ranking problem into several classification problems. The groundtruth pairwise depth ranking matrix $M_{i,j}$ for 3D pose $S_{3D} = ((x'_1, y'_1, z'_1), ..., (x'_n, y'_n, z'_n))$ is defined as follows:
\begin{eqnarray}\label{eq:M}
M_{ij} =
\begin{cases}
1   &  \text{if $z'_i > z'_j + \epsilon $,} \\
0   &  \text{if $z'_i < z'_j - \epsilon $,} \\
0.5   &  \text{if $|z'_i - z'_j| \leq \epsilon$.} \\
\end{cases}
\end{eqnarray}
Where $M_{ij}$ indicates the probability that the $i_{th}$ joint is behind $j_{th}$ joint and 
$\epsilon$ indicates the tolerable depth difference to avoid the ambiguity of two joints with very close depth. Thus predicting the depth ranking of $n$ joints is transformed to $n^2$ classification problems.

In the following paragraphs, we will introduce (1) how to learn pairwise depth ranking and (2) how to use pairwise depth ranking in predicting 3D joint positions. An overview of our framework is illustrated in Figure~\ref{fig:framework}. PRCNN predicts ranking matrix $M$ given the image $I$ and the 2D joint heat maps $H$. DPNet regresses 3D pose given the ranking matrix $M$ and the 2D joint locations $S_{2D}$.

\subsection{PRCNN: Learning Pairwise Rankings}
In order to estimate the ranking matrix, we propose the Pairwise Ranking Convolutional Neural Network, PRCNN. We concatenate generated 2D joint heatmaps $H_1, H_2, \ldots, H_N$ with the original image $I$ as the input of PRCNN. We adopt an 8-stack hourglass network \cite{newell2016stacked} as our 2D pose estimator. It is pretrained on MPII dataset \cite{andriluka20142d} and fine-tuned on Human3.6M \cite{ionescu2014human3}.

Inspired by RankNet \cite{burges2005learning} which ranks items pairwisely, the network PRCNN first extracts a one-dimensional feature of each joint, $F_1, ..., F_n$, then compute the difference between one-dimensional features of a joint pair as the feature of this joint pair. Residual network $\varGamma$ \cite{he2016deep} is used as the backbone of our feature extractor.
\begin{equation}
\varGamma(I, H_1, H_2, \ldots, H_N) = (F_1, F_2, \ldots, F_n)\in\mathbb{R}^n.
\end{equation}
\begin{equation}
F_{ij} \equiv F_i-F_j.
\end{equation}
Given the feature number of a joint pair, we apply the following rank transfer function to get the probability that $i_{th}$ joint has higher Z-value than $j_{th}$ joint, which means $i_{th}$ joint is behind $j_{th}$ joint.
\begin{equation}
P_{ij} = \frac{e^{F_{ij}}}{1+e^{F_{ij}}}.
\end{equation}
Note that $P_{ij}$ is in range $[0, 1]$ and becomes 0.5 when $i = j$. It is consistent with our representation, indicating the ranking probability of joint pairs.
Let ranking matrix $M$ be the desired target values. We adopt cross entropy loss as loss function which is frequently used in classification problems. For training, the probabilistic ranking cost function proposed in \cite{burges2005learning} is defined as:
\begin{equation}
\begin{aligned}
C_{ij} \equiv C(F_{ij}) & = -M_{ij}\log{P_{ij}} - (1-M_{ij})\log(1-P_{ij}) \\
& =-M_{ij}F_{ij}+\log(1+e^{F_{ij}}).
\end{aligned}
\end{equation}
The final cost function is the summation of all $C_{ij}$. Thus, this method turns the ranking task into several classification problems.
During inference, the output $P_{ij}$ is discretized into three values $0,1,0.5$ with corresponding intervals $[0, 0.5-\varepsilon),(0.5+\varepsilon, 1], [0.5-\varepsilon, 0.5+\epsilon]$ as the final output where $\varepsilon$ is a threshold.

Different from RankNet, where each feature depends on one item, PRCNN requires extracting all features from one image and predicts all of the pairwise rankings together. Hence we apply 19-channel tensors (3 for image and 16 for heatmaps) as inputs for our model.
We adopt Resnet-34 as the backbone of PRCNN and train it from scratch. We find that further increasing network depth doesn't improve the performance. Data augmentation for PRCNN is performed such as rotation, scaling and flipping like other 2D pose estimation methods. The network performs well on the Human3.6M dataset and gives reasonable results on the non-lab dataset such as MPII as well. Under protocol \#3 mentioned in section \ref{protocols}, which use three camera perspectives for training and another camera perspective for testing, the performance doesn't get worse. It proves that PRCNN can generalize well across different camera perspectives.

\subsection{DPNet: 3D Pose Estimation via Depth Ranking}
\begin{figure}[t]
	\setlength{\abovecaptionskip}{0.2cm}	
	\setlength{\belowcaptionskip}{-0.35cm}
	\centering
	\includegraphics[width = .5\textwidth]{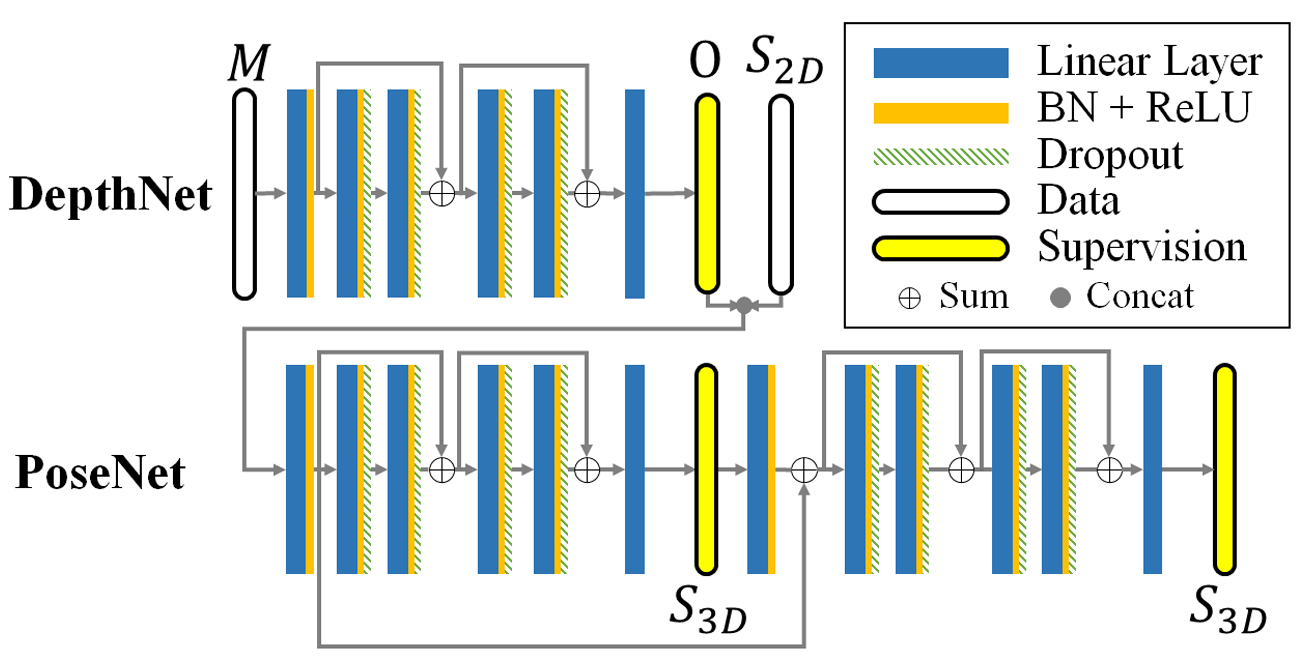}
	\caption{
		DPNet consists of DepthNet and PoseNet. DepthNet takes the pairwise ranking matrix $M$ as input and outputs the coarse depth $O$. PoseNet regresses 3D poses from $O$ and $S_{2D}$.}
	\label{fig_rpnet}
\end{figure}
According to previous paragraphs, with the knowledge of pairwise depth ranking, 2D pose $S_{2D}$ and human prior knowledge, the 3D human pose is almost determined geometrically. Thus we use only the predicted ranking matrix $P$, and 2D pose $S_{2D}$ as input at this stage.

However, unlike traditional geometric problem that has perfect input, despite a majority of the pairwise ranking matrix $P$ is correct, there is a portion of entries of $P$ provides noisy information. As shown in Figure~\ref{fig_ranking}, some elements in the ranking matrix are hard to learn if there is no clear evidence of the image. Directly learning $S_{3D}$ from the ranking matrix $P$ and $S_{2D}$ provides less accurate results. A coarse-to-fine network is proposed to resolve noisy information from the ranking matrix.

The first part of our coarse-to-fine network is called \textit{DepthNet}, which converts the ranking matrix $P$ into coarse depth values.
Given the input ranking matrix $P$, the DepthNet predicts coarse depth $O$ that is consistent with the ranking matrix.
The ground-truth of $O$ is the ranking order on Z-axis of each human joint. Before normalization, it is a permutation of $\{1,2, …, 16\}$ according to the depth values in $S_{3D}$.
By doing so, the network is trained to convert the noisy ranking matrix into coarse depth values. Thus noisy ranking pairs can be refined by majority correct ranking pairs. The refinement strategy generates more robust depth values than traditional methods such as topological sort algorithm.

The second part of our coarse-to-fine network \textit{PoseNet} combines coarse depth values with $S_{2D}$ and predicts more and more accurate 3D pose $S_{3D}.$  Inspired by the advance of multi-stage architectures and coarse-to-fine mechanisms \cite{pavlakos2017coarse,newell2016stacked},
we use a cascaded regression network with two stages. The first stage predicts the $S_{3D}$ directly and the second stage predicts its residuals. Each stage outputs are supervised by the 3D pose ground-truth $S_{3D}$. Within each stage, we employ two residual blocks following \cite{martinez2017simple}.
The architecture of our coarse-to-fine network is illustrated in Figure~\ref{fig_rpnet}. We use mean square error (MSE) to calculate the loss of each supervised layer and sum them up:
\begin{equation}
\mathcal{L}=\mathcal{L}_{O}+\mathcal{L}_{S_{3D}}+\mathcal{L}'_{S_{3D}}.
\end{equation}
where $\mathcal{L}_{O}$ is the loss of DepthNet, $\mathcal{L}_{S_{3D}}$ is the loss of the first stage of PoseNet, and $\mathcal{L}'_{S_{3D}}$ is for the second stage.
In order to remove the difference in magnitude of the variables and global shift. Both the supervised $S_{3D}$ and coarse depth $O$ are normalized to values whose mean value equals to $0$ and standard deviation equals to $1$.

\subsection{Data Augmentation}
As mentioned in \cite{fang2017learning}, data augmentation in 3D space is very effective for protocol \#3 experiment, whose training data are from three camera positions and test data are from another different camera positions. With depth ranking, data augmentation is still available and becomes more powerful on protocol \#3 experiment.

Data augmentation is performed by synthesizing input from virtual cameras. We synthesize the $S'_{3D}$ in virtual camera coordinate according to ground-truth 3D pose $S_{3D}$. Then, we project $S'_{3D}$ on camera perspective plane to obtain 2D joint locations $S'_{2D}$ and compute ranking matrix $P'$ concerning $S'_{3D}$. Currently, augmented data are generated from ground truth $S_{3D}$. However, the estimated ranking matrix from PRCNN and 2D joint locations from 2D pose estimator can be noisy.
Thus we use Gaussian mixture model (GMM) mentioned in \cite{fang2017learning} to add noise to $S'_{2D}$.
We also propose a statistical method to adjust each entry of the pairwise ranking matrix $P'_{ij}$ based on its accuracy.
We calculate the accuracy of each entry $p_{ij}$ according to the predicted ranking matrix $P$ in training set. We flip the synthesized $P'_{ij}$ with the probability of $1-p_{ij}$.

Previous work \cite{fang2017learning} samples virtual cameras on the same circle where three cameras from training data lie. The assumption and prior knowledge about camera settings used in data augmentation are strong. However, our methods only assume that all cameras roughly point towards the performer with motion capture device. We first calculate the rough position of the performer by finding the center position $V_c$ that is the closest point to optical axes of all cameras.
\begin{equation}\label{opti}
V_c = \argmin_{V} \sum_{i=1}^{k} d(V,l_i).
\end{equation}
where $l_i$ is the line indicating the optical axis. Then the distance $d$ between cameras to the center position $V_c$ are sampled using the normal distribution, whose center and variance are computed from training data. Then we sample the camera positions uniformly on the surface of the sphere with center $V_c$ and radius $d$. The optical axis is the line connecting the sampled camera position and $V_c$. One axis of the camera coordinate is parallel to the ground plane to make the synthesized $S'_{2D}$ upright.

\section{Experiments}
In this section, we introduce datasets and protocols first, then provide details how we implement our framework.
We evaluate our method on Human3.6M and compare with state-of-the-arts methods.
To verify impacts of each component in our approach, we also perform ablation studies.
Finally, the qualitative results visualize the 3D pose estimation results on Human3.6M and MPII dataset.

\subsection{Datasets and Protocols}\label{protocols}
\textbf{Human3.6M} is currently the largest public 3D human pose benchmark.
The dataset captured human poses in a laboratory environment with Motion Capture technology.
It consists of 3.6 million images describing daily activities.
There are 4 cameras, 11 subjects (actors) and 17 scenarios (actions) in this dataset.
We use mean per joint position error (MPJPE) as evaluation metric and adopt it in three protocols described in previous works \cite{fang2017learning}.
\begin{itemize}[leftmargin=*]
	\item
	\textit{Protocol \#1} uses subjects S1, S5, S6, S7, S8 for training and S9 and S11 for testing.
	It is a widely used benchmark when using Human3.6M dataset.
	\item
	\textit{Protocol \#2} is based on \textit{Protocol \#1} and aligns the estimated 3D pose to the groundtruth by a rigid transformation called Procrustes Analysis, which is the protocol to evaluate the correctness of the 3D pose structure.
	\item
	\textit{Protocol \#3} aims to evaluate the generality of methods on camera parameters and uses 3 cameras views for training and the other one for testing\cite{fang2017learning}.
\end{itemize}
\textbf{MPII} is widely used for 2D human pose estimation in the wild. We will provide qualitative evaluation on this dataset.

\subsection{Implementation Details}

\textbf{2D pose estimation}
We follow the configurations in \cite{martinez2017simple} and use the stacked hourglass \cite{newell2016stacked} as 2D pose estimator. The variance of Gaussian is set to 4 in our experiments.

\textbf{PRCNN}
The PRCNN is based on the Deep Residual Network.
Differently, we set the size of the last full connected layer to $N=16$. We implement the pairwise layer and ranking transfer layer of RankNet \cite{burges2005learning} to obtain ranking matrix.
We train the PRCNN model with binary cross entropy loss and use Stochastic Gradient Descent (SGD) to train 25 epochs over the whole training set. In all experiments, the models are trained on 8 TITAN Xp GPUs with batch size 64 and the initial learning rate 0.1.

\textbf{DPNet}
We set the root of 3D pose to (0,0,0) following \cite{martinez2017simple}.
We train our DPNet for 400 epochs using Adam, and the initial learning rate is 0.001 with exponential decay. The mini-batches is set to 64.
The probability of dropout is set to 0.3 so that it remains more possible information in rankings.
With the benefit of low dimensionality, we only use one TITAN Xp GPU to train this network.
In protocol \#3, the scale of the augmented dataset is triple as the original Human3.6M dataset.

\subsection{Comparisons with State-of-the-art Methods}
We compare the proposed method with the state-of-the-art methods on Human 3.6M dataset. The comparisons with both end-to-end and two-stage methods under all protocols are shown in Table \ref{human3.6M}. There are three observations as follows:
(1) The depth ranking is an efficient feature, and the proposed DRPose3D outperforms the state-of-the-art method including both end-to-end \cite{Sun_2017_ICCV,zhou2017towards,zhou2017monocap,zhou2016deep} and two-stage methods \cite{martinez2017simple,fang2017learning} under all protocols. 
(2) Augmentation is effect \cite{martinez2017simple,fang2017learning}. 
(3) Depth ranking improves the robustness of DPNet. The proposed method achieves a reconstruct error of $69.0mm$ without augmentation on protocol \#3, $3.8mm$ errors drop from \cite{fang2017learning}. Observed results show that depth rankings eliminate the ambiguities, so that prevent our network from overfitting to specified camera perspectives.

\begin{figure}[tbh]
	\setlength{\abovecaptionskip}{0.cm}	
	\setlength{\belowcaptionskip}{0cm}
	\centering
	\includegraphics[width = .5\textwidth]{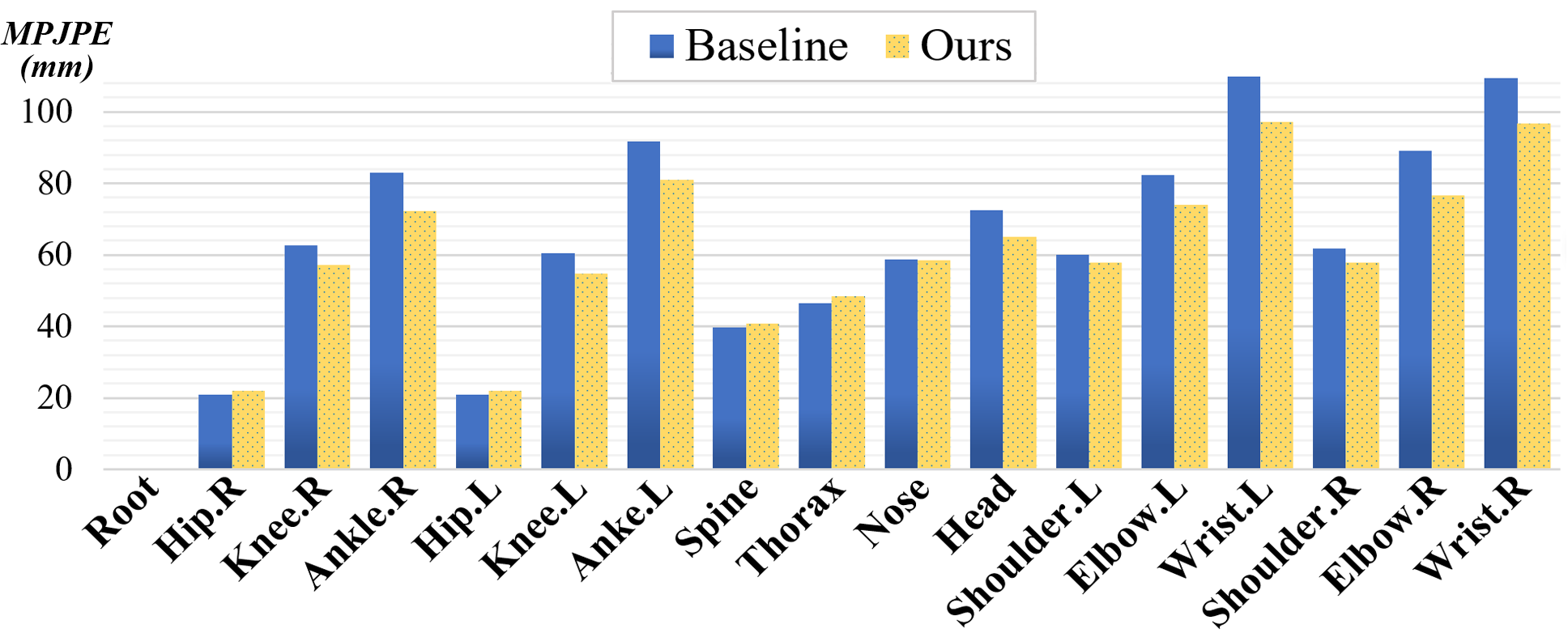}
	\caption{Comparisons in joint level.}
	\label{fig_errors}
\end{figure}

To verify that the depth ranking improves the baseline network, we evaluate all joints by MPJPE.
The result shown in Figure \ref{fig_errors} indicates that
the joints with lager reachable workspace like wrists, elbows, and feet can provide more robust ranking cues and obtain larger improvements than other joints. For example, the right wrist achieves $12.53\%$ ($109.54mm\rightarrow97.01mm$) errors drop while thorax only achieves $2.36\%$ ($46.40mm\rightarrow45.45mm$).

\subsection{Upper Bound of Our Approach}
\begin{table}[tb]
	\setlength{\abovecaptionskip}{0.2cm}
	\setlength{\belowcaptionskip}{-0.2cm}
	\centering
	\small
	\begin{tabular}{ccc}
		\specialrule{1pt}{1pt}{1pt}
		2D Pose & Depth Ranking & MPJPE($mm$) \\
		\specialrule{0.5pt}{1pt}{2pt}
		GT      & None     & 45.5  \\
		GT      & Predicted  & 41.2  \\
		GT      & GT       & 30.2 \\
		\specialrule{1pt}{1pt}{1pt}
	\end{tabular}
	\caption{Evaluation under Protocol \#1 with estimation based on groundtruth or predicted inputs.}
	\label{gt}
\end{table}

To demonstrate our statement that depth ranking improves the network, we use ground-truth 2D poses and pairwise ranking matrix to explore the upper bound of our approach, as shown in Table~\ref{gt}.
The result shows that: (1) After using ranking matrix predicted by PRCNN, we get $4.3mm$ errors drop on MPJPE metric. (2) By using the ground truth 2D poses and depth rankings, the proposed method achieves $30.2mm$ on Human 3.6M. It is significantly promoted by $33.63\%$ compared to only performing ground truth 2D positions. (3) It proves that more accurate ranking estimation could further improve the 3D pose estimation.

\subsection{Ablation Study}
We study each component of our method on the Human3.6M dataset to verify the effectiveness.

\begin{figure}[tb]
	\setlength{\abovecaptionskip}{0.2cm}	
	\setlength{\belowcaptionskip}{-0.2cm}
	\centering
	\includegraphics[width = .5\textwidth]{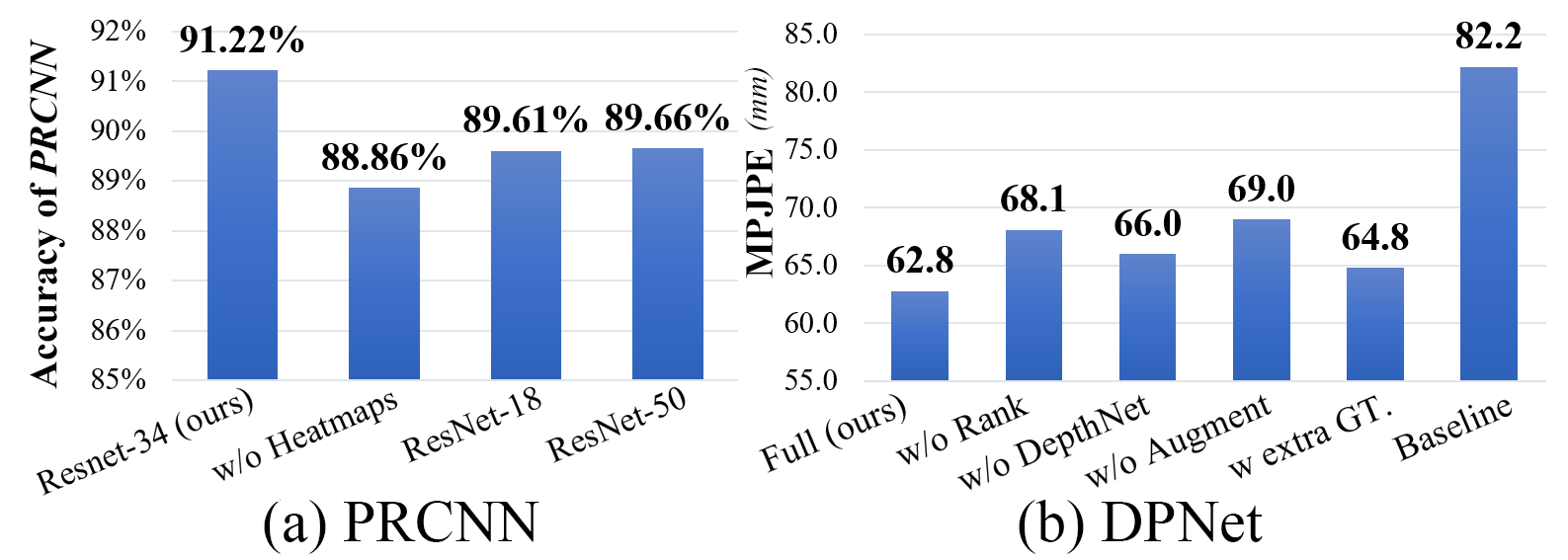}
	\caption{(a) The accuracy of ranking matrix with difference designs. (b) Ablative analysis of DPNet}
	\label{fig_ablation}
\end{figure}

\begin{table*}[t]
	\setlength{\abovecaptionskip}{0.1cm}
	\setlength{\belowcaptionskip}{-0.5cm}
	\centering
	\small
	\setlength{\intextsep}{0cm}
	\setlength{\textfloatsep}{0pt}
	\resizebox{
		1.\textwidth}{!}{
		\begin{tabular}{@{}lcccccccccccccccc@{}}
			\specialrule{1pt}{1pt}{2pt}
			Protocol \#1 & Direction & Discuss & Eat & Greet & Phone & Photo & Pose & Purchase & Sit & SitDown & Smoke & Wait & WalkDog & Walk & WalkT. & Avg.\\
			\specialrule{0.5pt}{0pt}{1pt}
			LinKDE (PAMI'16)           & 132.7         & 183.6         & 132.3         & 164.4         & 162.1       & 205.9         & 150.6         & 171.3         & 151.6         & 243.0         & 162.1         & 170.7         & 177.1         & 96.6          & 127.9         & 162.1         \\
			Zhou et al. (ECCV'16)      & 91.8          & 102.4         & 96.7          & 98.8          & 113.4       & 125.2         & 90.0          & 93.8          & 132.2         & 159.0         & 107.0         & 94.4          & 126.0         & 79.0          & 99.0          & 107.3         \\
			Pavlakos et al. (CVPR'17)  & 67.4          & 71.9          & 66.7          & 69.1          & 72.0        & 77.0          & 65.0          & 68.3          & 83.7          & 96.5          & 71.1          & 65.8          & 74.9          & 59.1          & 63.2          & 71.9          \\
			Zhou et al. (ICCV'17)      & 54.8          & 60.7          & 58.2          & 71.4          & 62.0 & \textbf{65.5} & 53.8          & 55.6          & 75.2          & 111.6         & 64.1          & 66.0          & \textbf{51.4}          & 63.2          & 55.3          & 64.9          \\
			Martinez et al. (ICCV'17)  & 51.8          & 56.2          & 58.1          & 59.0          & 69.5        & 78.4          & 55.2          & 58.1          & 74.0          & 94.6          & 62.3          & 59.1          & 65.1          & 49.5          & 52.4          & 62.9          \\
			Fang et al. (AAAI'18)      & 50.1          & \textbf{54.3}          & 57.0          & 57.1          & 66.6        & 73.3          & 53.4          & 55.7          & 72.8          & 88.6          & 60.3          & 57.7          & 62.7          & 47.5          & 50.6          & 60.4          \\
			Sun et al. (ICCV'17)      & 52.8      & 54.8    & 54.2   & 54.3  & \textbf{61.8}  & 67.2  & 53.1 & 53.6     & 71.7    & 86.7        & 61.5  & 53.4 & 61.6   & 47.1 & 53.4   & 59.1 \\
			\specialrule{0.5pt}{0pt}{2pt}
			Ours                       & \textbf{49.2} & 55.5 & \textbf{53.6} & \textbf{53.4} & 63.8        & 67.7          & \textbf{50.2} & \textbf{51.9} & \textbf{70.3} & \textbf{81.5} & \textbf{57.7} & \textbf{51.5} & 58.6 & \textbf{44.6} & \textbf{47.2} & \textbf{57.8} \\
			\specialrule{1pt}{1pt}{2pt}
			Protocol \#2                & Direction & Discuss & Eat & Greet & Phone & Photo & Pose & Purchase & Sit & SitDown & Smoke & Wait & WalkDog & Walk & WalkT. & Avg. \\
			\specialrule{0.5pt}{0pt}{1pt}
			Bogo et al. (ECCV'16)     & 62.0      & 60.2    & 67.8   & 76.5  & 92.1  & 77.0  & 73.0 & 75.3     & 100.3   & 137.3       & 83.4  & 77.3 & 86.8   & 79.7 & 87.7   & 82.3 \\
			Moreno-Noguer (CVPR'17)   & 66.1      & 61.7    & 84.5   & 73.7  & 65.2  & 67.2  & 60.9 & 67.3     & 103.5   & 74.6        & 92.6  & 69.6 & 71.5   & 78.0 & 73.2   & 74.0 \\
			Zhou et al. (Arxiv'17)    & 47.9      & 48.8    & 52.7   & 55.0  & 56.8  & 65.5  & 49.0 & 45.5     & 60.8    & 81.1        & 53.7  & 51.6 & 54.8   & 50.4 & 55.9   & 55.3 \\
			Sun et al. (ICCV'17)      & 42.1      & 44.3    & 45.0   & 45.4  & 51.5  & 53.0  & 43.2 & 41.3     & 59.3    & 73.3        & 51.0  & 44.0 & 48.0   & 38.3 & 44.8   & 48.3 \\
			Martinez et al. (ICCV'17) & 39.5      & 43.2    & 46.4   & 47.0  & 51.0  & 56.0  & 41.4 & 40.6     & 56.5    & 69.4        & 49.2  & 45.0 & 49.5   & 38.0 & 43.1   & 47.7 \\
			Fang et al. (AAAI'18)     & 38.2      & 41.7    & 43.7   & 44.9  & 48.5  & 55.3  & 40.2 & 38.2     & 54.5    & 64.4        & 47.2  & 44.3 & 47.3   & 36.7 & 41.7   & 45.7 \\
			\specialrule{0.5pt}{0pt}{2pt}
			Ours                      & \textbf{36.6}      & \textbf{41.0}    & \textbf{40.8}   & \textbf{41.7}  & \textbf{45.9}  & \textbf{48.0}  & \textbf{37.0} & \textbf{37.1}     & \textbf{51.9}    & \textbf{60.4}        & \textbf{43.9}  & \textbf{38.4} & \textbf{42.7}     & \textbf{32.9} & \textbf{37.2}   & \textbf{42.9} \\
			\specialrule{1pt}{1pt}{2pt}
			Protocol \#3                & Direction     & Discuss       & Eat           & Greet         & Phone         & Photo         & Pose          & Purchase      & Sit           & SitDown        & Smoke         & Wait          & WalkDog       & Walk        & WalkT.        & Avg.        \\
			\specialrule{0.5pt}{0pt}{1pt}
			Pavlakos et al. (CVPR'17) & 79.2          & 85.2          & 78.3          & 89.9          & 86.3          & 87.9          & 75.8          & 81.8          & 106.4         & 137.6          & 86.2          & 92.3          & 72.9          & 82.3        & 77.5          & 88.6        \\
			Martinez et al. (ICCV'17) & 65.7          & 68.8          & 92.6          & 79.9          & 84.5          & 100.4         & 72.3          & 88.2          & 109.5         & 130.8          & 76.9          & 81.4          & 85.5          & 69.1        & 68.2          & 84.9        \\
			Zhou et al. (ICCV'17)     & 61.4          & 70.7          & 62.2          & 76.9          & 71.0          & 81.2          & 67.3          & 71.6          & 96.7          & 126.1          & 68.1          & 76.7          & 63.3          & 72.1        & 68.9          & 75.6        \\
			Fang et al. (AAAI'18)     & 57.5          & 57.8          & 81.6          & 68.8          & 75.1          & 85.8          & 61.6          & 70.4          & 95.8          & 106.9 & 68.5          & 70.4          & 73.89         & 58.5        & 59.6          & 72.8        \\
			\specialrule{0.5pt}{0pt}{1pt}
			Ours w/o augmentation	& \textbf{53.6}          & 56.5          & 73.2          & 66.6          & 72.8          & 79.6          & 56.4          & 71.1          & 87.4          & 106.3 & 65.2          & 64.3          & 69.7         & 58.8        & 57.5          & 69.0        \\
			Ours                      & 55.8 & \textbf{56.1} & \textbf{59.0} & \textbf{59.3} & \textbf{66.8} & \textbf{70.9} & \textbf{54.0} & \textbf{55.0} & \textbf{78.8} & \textbf{92.4}          & \textbf{58.9} & \textbf{56.2} & \textbf{64.6} & \textbf{56.6} & \textbf{55.5} & \textbf{62.8} \\
			\specialrule{1pt}{1pt}{2pt}
		\end{tabular}
	}
		\caption{Quantitative comparisons of Mean Per Joint Position Error ($mm$) between the estimated pose and the ground-truth on Human3.6M under Protocol \#1, Protocol \#2 and Protocol \#3. The best score is marked in bold.}
		\label{human3.6M}
\end{table*}
%
We perform an ablative analysis to understand the impact of the design choices of our PRCNN.
We present the results in Figure~\ref{fig_ablation}~(a).
When taking only original images as inputs, the model(w/o heatmaps) achieves the accuracy of $88.86\%$.
Combing heatmaps of human joints and original image increases the accuracy by $2.36\%$, which proves the effectiveness of combining semantic knowledge explicitly.
We have also tried basic network with different depth, i.e., Resnet-18, Resnet-34, and Resnet-50. 
Since we only train PRCNN on Human3.6M dataset, whose images have almost the same backgrounds, very deep network like Resnet-50 may cause overfitting.
Resnet-34 achieves $91.22\%$ mean accuracy in protocol \#1 and is chosen for all the other experiments.

\begin{figure}[t]
	\setlength{\abovecaptionskip}{0.1cm}
	\setlength{\belowcaptionskip}{-0.4cm}
	\setlength{\intextsep}{0cm}
	\centering
	\includegraphics[width = .45\textwidth]{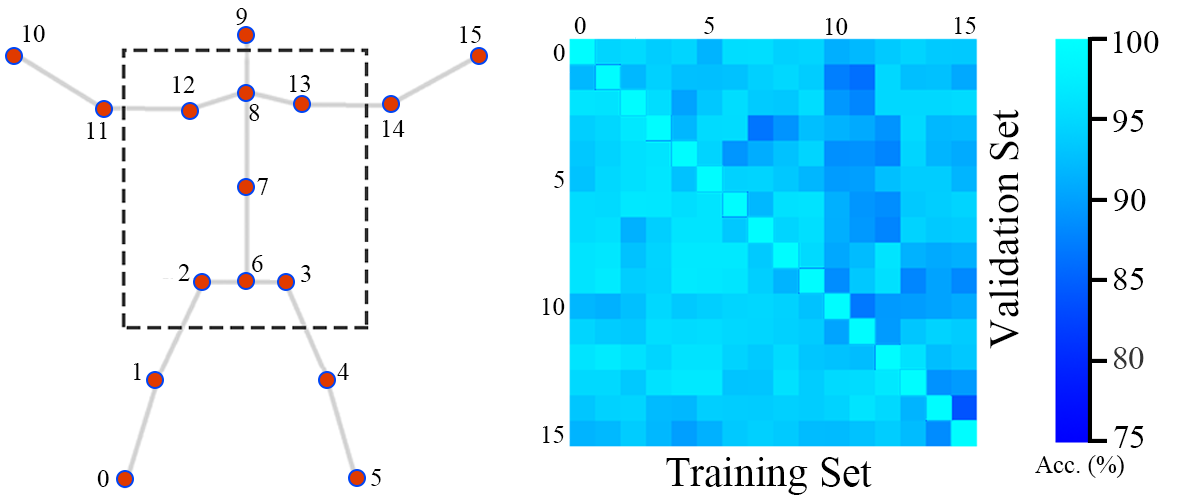}
	\caption{Results of pairwise ranking predictions. The rankings of joint pairs inside the dashed rectangle are less accurate.}
	\label{fig_ranking}
\end{figure}

We further illustrate pairwise rankings accuracy in Figure \ref{fig_ranking}.
Bright block in the ranking matrix indicates high accuracy.
We find that connections between joints in the torso, inside dashed line box such as \textit{Hip.R-Spine}, have lower accuracy because their depths are too close to be recognized.
However, relations like right and left shoulders with 96.71\% accuracy are accessible to indicate which direction the subject is facing.

Figure~\ref{fig_ablation}(b) shows component analysis of DPNet.
To evaluate the cross-camera-perspective effect, these experiments are conducted under protocol \#3.
Our proposed method with all components achieves the result of $62.8mm$ and exceeds the baseline (w/o Rank\&Augment) by a large margin ($19.4mm\downarrow$).
When we remove the depth ranking, MPJPE increases to $68.1mm$, showing that depth rankings effectively enhance the regression from 2D to 3D pose.
The model without DepthNet, directly combing 2D joint locations with the noisy ranking matrix, leads to a growth of $3.2mm$ errors, indicating that DepthNet can reduce the noise in pairwise ranking matrix effectively.
After that, we evaluate the effectiveness of data augmentation.
The result shows augmentation with ground-truth ranking matrix and 2D joint locations generated by virtual cameras achieves better performance ($64.8mm$) than the model without augmentation ($69.0mm$). The statistical augmentation method for ranking matrix further decreases the joint error by $2.1mm$.

\subsection{Qualitative Results}

\begin{figure}[b]
	\setlength{\abovecaptionskip}{0.cm}	
	\setlength{\belowcaptionskip}{-0.4cm}
	\setlength{\intextsep}{-0.4cm}
	\centering
	\includegraphics[width = 0.43\textwidth]{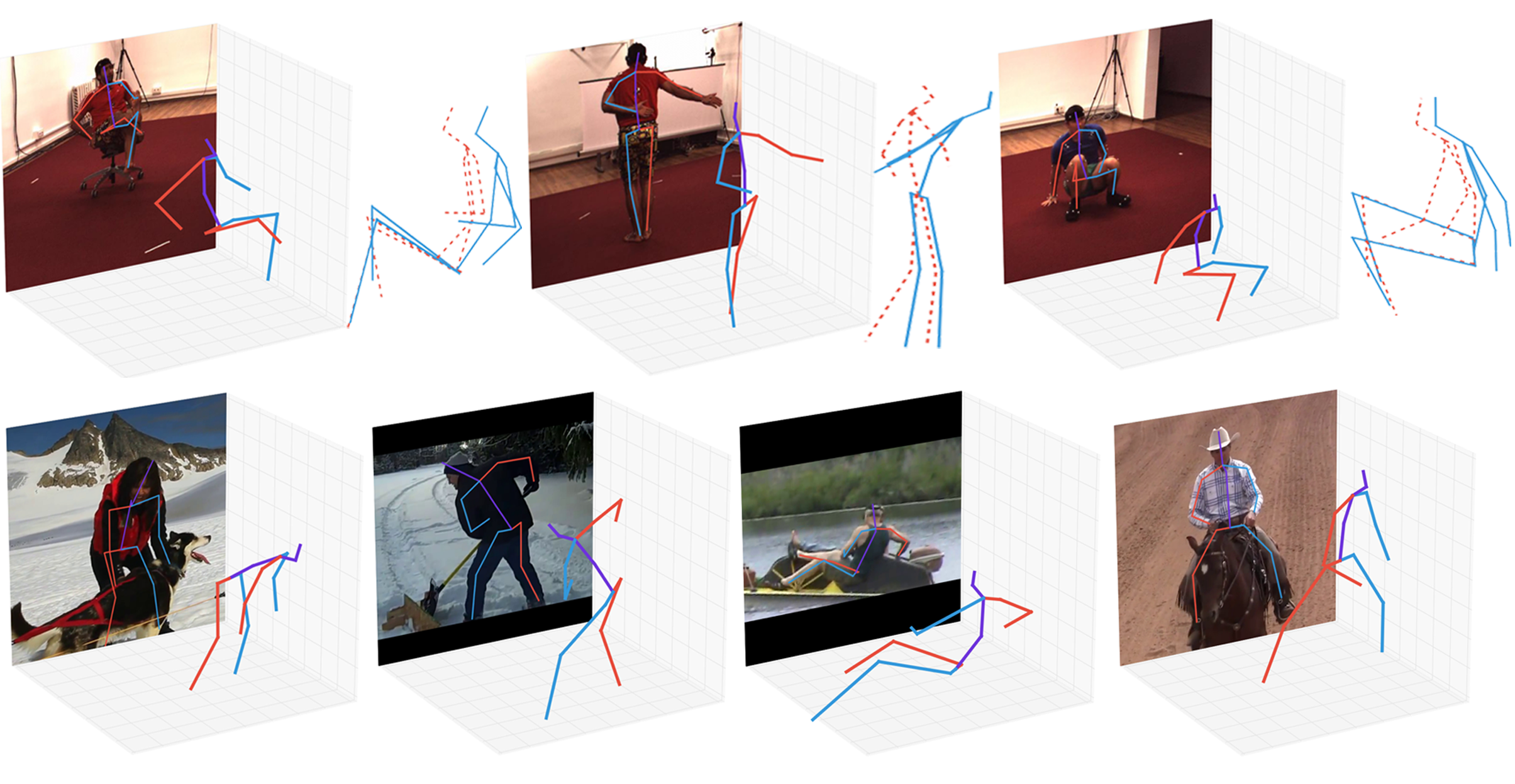}
	\caption{
		Qualitative Results. Top: DRPose3D (blue) and the baseline method (red dashed) on Human3.6M. Bottom: MPII results. }
	\label{fig:qualitative}
\end{figure}

Since our DPNet is trained based on 2D locations and rankings, it is possible to estimate 3D poses with images in the wild. We give qualitative results on both Human3.6M dataset and MPII dataset in Figure~\ref{fig:qualitative}.
The first row illustrated some examples from Human3.6M.
The red-dotted line is the baseline estimation while blue line indicates ours.
Depth ranking provides geometric knowledge that eliminates the ambiguities of the limbs and corrects the angle of the torso as shown in the side view of the top row samples.

More challenging samples from MPII dataset are shown in the bottom row. The proposed DRPose3D presents to have better generality and can obtain reasonable 3D poses even in some challenging cases: the subject lies down or do exaggerated actions.

\section{Conclusion}
In this paper, we propose a two-stage DRPose3D approach to tackle the 3D pose estimation task. The proposed method involves depth ranking to fully utilize 2D human pose datasets and 3D information contained in images. To extract the depth ranking from a single RGB image, we first design a PRCNN model to generate pair-wise depth relation between human joints. After that, a coarse-to-fine 3D pose estimator is proposed to predict 3D human pose from both 2D joint and depth ranking. Finally, we explore data augmentation for DRPose3D and prove that depth ranking can further enlarge improvement brought by data augmentation.
Overall, the proposed method achieves the state-of-the-art results on three common protocols of the Human3.6M dataset.

\section*{Acknowledgments}
This work is supported by National Natural Science Foundation of China (No. 61472245), and the Science and Technology Commission of Shanghai Municipality Program (No. 16511101300).

\appendix

\bibliographystyle{named}
\bibliography{drpose3D}

\end{document}